\documentclass[sigconf]{acmart}
\AtBeginDocument{%
  }

\copyrightyear{2025}
\acmYear{2025}
\setcopyright{acmlicensed}\acmConference[WWW Companion '25]{Companion Proceedings of the ACM Web Conference 2025}{April 28-May 2, 2025}{Sydney, NSW, Australia}
\acmBooktitle{Companion Proceedings of the ACM Web Conference 2025 (WWW Companion '25), April 28-May 2, 2025, Sydney, NSW, Australia}
\acmDOI{10.1145/3701716.3715501}
\acmISBN{979-8-4007-1331-6/25/04}

\usepackage{balance}
\usepackage{bbding}

\begin{document}

\title[CAIRNS: Balancing Readability and Accuracy]{CAIRNS: Balancing Readability and Scientific Accuracy in Climate Adaptation Question Answering}


\author{Liangji Kong}
\email{liangji.kong@student.unsw.edu.au}
\affiliation{%
  \institution{University of New South Wales}
  \city{Sydney}
  \country{Australia}}

\author{Aditya Joshi}
\email{aditya.joshi@unsw.edu.au}
\affiliation{%
  \institution{University of New South Wales}
  \city{Sydney}
  \country{Australia}}

\author{Sarvnaz Karimi}
\email{sarvnaz.karimi@csiro.au}
\affiliation{%
  \institution{CSIRO's Data61}
  \city{Sydney}
  \country{Australia}}

\renewcommand{\shortauthors}{Kong, Joshi, Karimi}
\renewcommand{\shortauthors}{Kong et al.}

\begin{abstract}

Climate adaptation strategies are proposed in response to climate change. They are practised in agriculture to sustain food production. These strategies can be found in unstructured data (for example, scientific literature from the Elsevier website) or structured (heterogeneous climate data via government APIs). We present Climate Adaptation question-answering with Improved Readability and Noted Sources (CAIRNS), a framework that enables experts---farmer advisors---to obtain credible preliminary answers from complex evidence sources from the web. It enhances readability and citation reliability through a structured ScholarGuide prompt and achieves robust evaluation via a consistency-weighted hybrid evaluator that leverages inter-model agreement with experts. Together, these components enable readable, verifiable, and domain-grounded question-answering without fine-tuning or reinforcement learning. Using a previously reported dataset of expert-curated question-answers, we show that CAIRNS outperforms the baselines on most of the metrics. Our thorough ablation study confirms the results on all metrics. To validate our LLM-based evaluation, we also report an analysis of correlations against human judgment.
\end{abstract}



\received{17 November 2025}
\received[revised]{xxx}
\received[accepted]{xxx}

\maketitle


\section{Introduction}
\label{sec:introduction}
Climate adaptation deals with supporting people and ecosystems in the wake of projected climate impacts~\cite{runhaar2018mainstreaming}. In agriculture, the knowledge of adaptation strategies is written in the scientific literature and industry reports. One way to make this knowledge accessible is through  Question-Answering (QA) tools, which could utilise large language models (LLMs) with retrieval-augmented generation (RAG) from online data sources~\cite{Nguyen2025ClimateAdaptationQA}. A previously reported RAG system for climate adaptation QA in agriculture identifies three critical gaps: (1) localisation of queries; (2) integration of evidence from multiple sources, such as climate data APIs, scientific literature or industry reports; and (3) evaluation of these systems with experts~\cite{Nguyen2025MyClimateCoPilot}.  Moreover, these systems often face a trade-off between factual accuracy and human readability because scientifically rigorous answers tend to be less accessible, whereas fluent narratives risk oversimplifying or distorting evidence. In this paper, we address the three gaps and propose a trade-off mechanism called   \textbf{C}limate \textbf{A}daptation question-answering with \textbf{I}mproved \textbf{R}eadability and \textbf{N}oted \textbf{S}ources (\textbf{CAIRNS}), a \textit{location-aware and evidence-traceable}, a QA framework for agricultural climate adaptation. This work is aligned with the United Nations Sustainable Development Goal (SDG) 13, \textit{Climate Action}, aiming to balance these competing goals by combining structured prompting for readability with strict evidence constraints for factual integrity. CAIRNS first identifies and normalizes location entities, then applies location-weighted retrieval to align regional relevance. For evaluation, CAIRNS adopts a hybrid framework that combines a seven-dimensional evaluation rubric with faithfulness metrics. Using an expert-curated dataset of 50 climate adaptation question-answers that reflect real information needs and requiring domain knowledge to answer, CAIRNS reports an improved specificity and verifiability, but also achieves a better trade-off between location prompting and retrieval precision, as compared with baselines and its ablations. Our contributions are: (1) a unified framework integrating online research papers and climate data that results in state-of-the-art average results; (2) ScholarGuide prompt, a structured prompting technique balancing readability and accuracy; and (3) a human-LLM correlation analysis to solidify the claims.

\section{Related Work}

Retrieval-augmented generation (RAG) has significantly improved large language models on knowledge-intensive tasks~\cite{lewis2020retrieval}. However, most existing systems focus on generic or news-based domains, with limited optimization for high-stakes applications such as agricultural climate adaptation. Recent domain-specific QA studies~\cite{Nguyen2025ClimateAdaptationQA, Nguyen2025MyClimateCoPilot, opoku2025rag} incorporate document retrieval and climate data, but still rely on static retrieval and generic prompting, limiting region-specific and verifiable answers. Geography-aware QA has been explored in other domains~\cite{chen2021geoqa, li2023location}, but these methods are not tailored to climate reasoning. Moreover, existing studies still treat climate data and literature as separate sources, lacking a unified evidence chain~\cite{Nguyen2025ClimateAdaptationQA, Nguyen2025MyClimateCoPilot}. On the evaluation side,  \citet{Nguyen2025ClimateAdaptationQA} proposed a seven-dimensional human rubric and examined consistency between LLM and expert assessments, establishing a foundation for human-aligned evaluation in this domain. Building on their rubric, CAIRNS introduces faithfulness metrics and a $\kappa$-weighted hybrid evaluator to align human and automated scoring for scalable domain QA, while its structured prompting component directly tackles the persistent trade-off between readability and factual accuracy often overlooked in prior RAG-based systems.

\section{Methodology}
\label{sec:methodology}

\begin{figure}[tb]
    \centering
    \includegraphics[width=0.9\linewidth]{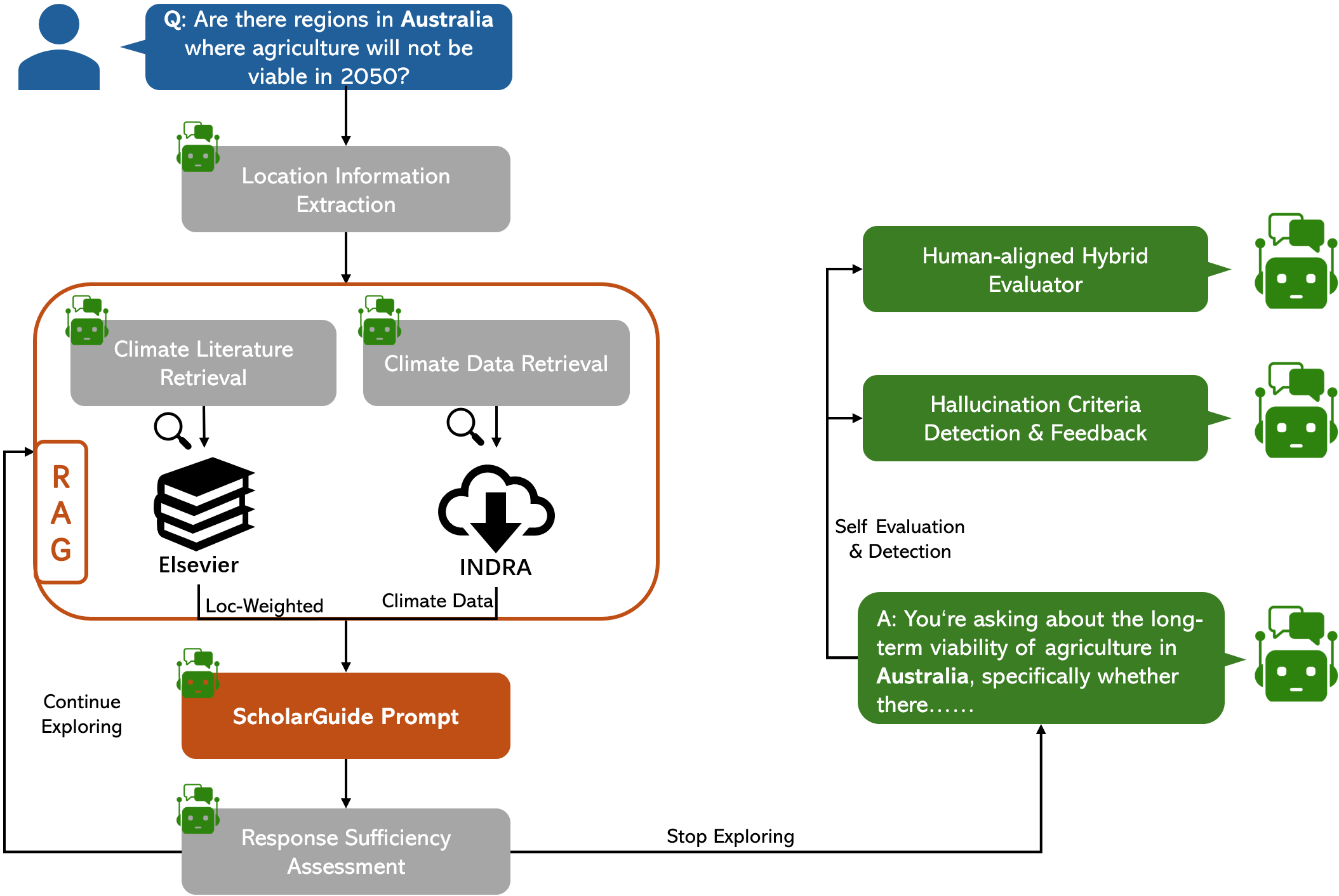}
    \caption{Overview of the CAIRNS framework.}
    \Description[<A flow diagram showing how CAIRNS retrieves climate literature and data, applies the ScholarGuide Prompt for reasoning, and performs self-evaluation to generate verifiable answers about agricultural adaptation questions.>]{<This figure illustrates the end-to-end process of the CAIRNS framework for trustworthy question answering in agricultural climate adaptation. A user asks a domain-specific question (e.g., “What are the ideal pollination conditions for growing almonds?”). The system first retrieves relevant climate literature using a location-weighted retrieval module, then retrieves structured climate data via the INDRA API. Both sources are fed into the ScholarGuide Prompt, which guides the model to construct a unified evidence chain and to reason step-by-step. The model then performs self-evaluation and hallucination detection based on epistemological, presentational, and citation criteria. If the response lacks sufficient evidence, the system re-enters the retrieval–generation loop; otherwise, it produces a final, citation-grounded answer. This workflow demonstrates how CAIRNS integrates retrieval, structured prompting, and self-assessment to improve answer credibility and geographic specificity.>}
    \label{fig:CAIRNS_pipeline}
\end{figure}

CAIRNS comprises four stages: (1) location-weighted retrieval on climate literature from the Elsevier website and data collection from a climate data API; (2) answer generation with ScholarGuide Prompt; (3) answer verification; and (4) hybrid evaluation. These stages (Figure~\ref{fig:CAIRNS_pipeline}) are described as follows.

To retrieve the relevant documents, we design a hybrid retriever that combines sparse lexical matching (Okapi BM25) and dense semantic retrieval ({\tt BAAI/bge-base-en-v1.5}~\cite{bge_embedding}) through a weighted scoring scheme. Hybrid retrieval alone is insufficient because location cues strongly affect relevance in climate adaptation QA. Therefore, we augment the base hybrid retrieval score with a location similarity term to balance semantic and spatial relevance:
\begin{equation}
\text{HybridScore}'=(1-\beta)\cdot\text{HybridScore}+\beta\cdot\text{LocSim}, \quad \beta\in[0,1].
\label{eq:location-weight}
\end{equation}
The location similarity (\text{LocSim}) measures overlap between the question’s location entities and those in each document. After retrieval, a two-stage \textit{action--parameter} reasoning process selects the appropriate climate-data API and generates parameters. The structured climate information returned by the API is converted into citation-ready text using verbalisation (for example, ``the average rainfall in Picton is 835 mm''), and merged with web-sourced literature references to form a unified, verifiable evidence chain.

\begin{figure}[tb]
    \centering
    \includegraphics[width=0.9\linewidth]{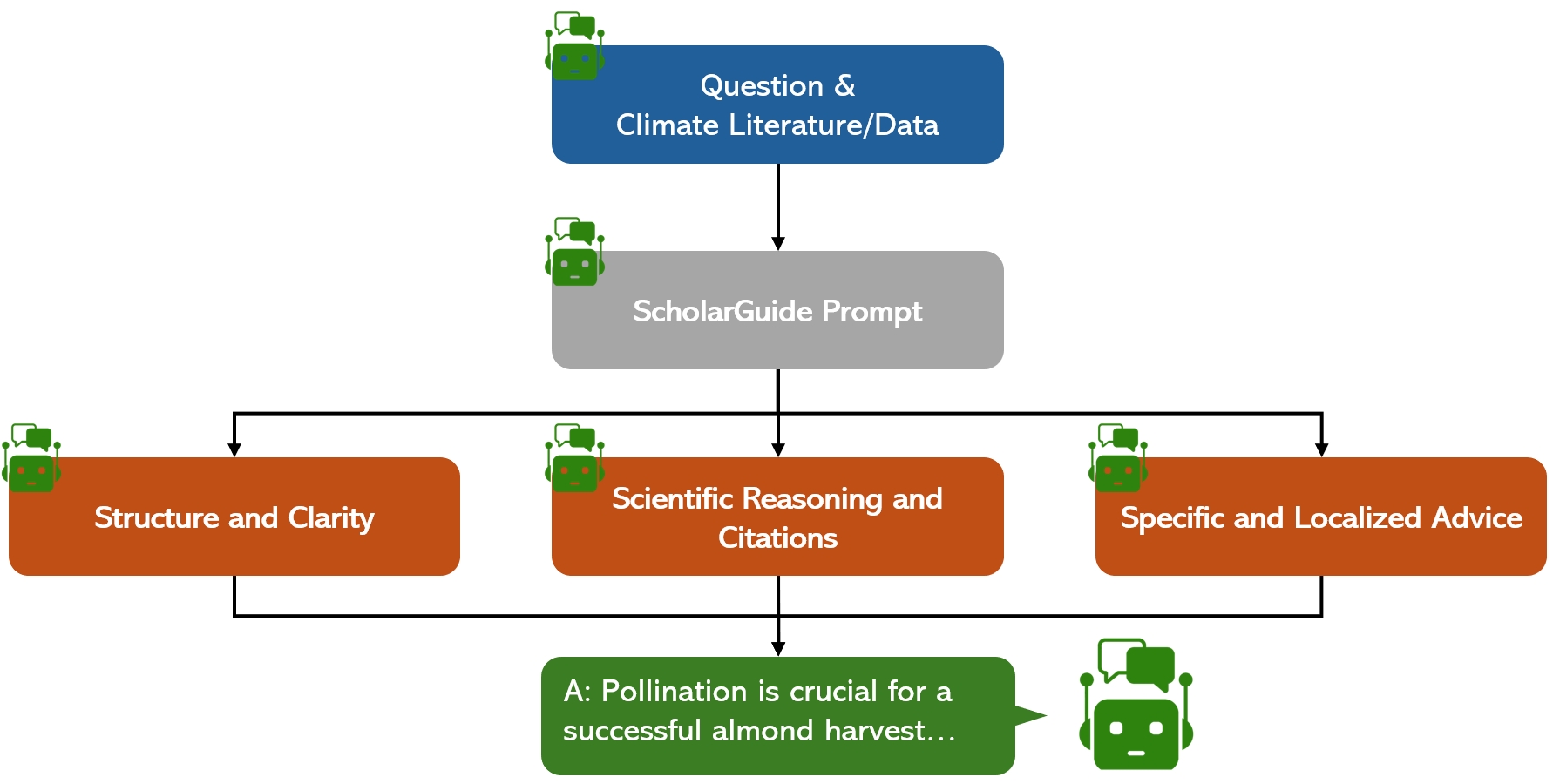}
    \caption{ScholarGuide Prompt: structured reasoning for readable and verifiable answers.}
    \Description[<A diagram illustrating how the ScholarGuide Prompt organizes a question and retrieved climate evidence into three reasoning layers—clarity, scientific citation, and localized specificity—to produce a well-structured, verifiable answer.>]{<This figure visualizes the internal workflow of the ScholarGuide Prompt (SGP). The process begins with a user question and the retrieved climate literature or data as inputs. The SGP structures the generation process into three reasoning layers: (1) Structure and Clarity—ensuring logical flow and well-organized presentation; (2) Scientific Reasoning and Citations—requiring explicit scientific justification and proper reference to supporting literature; and (3) Specific and Localized Advice—adapting the answer to the geographic and agricultural context of the query. The three layers jointly refine the model’s response into a final, human-readable, and citation-grounded answer (e.g., “Pollination is crucial for a successful almond harvest…”), illustrating how SGP enforces both readability and verifiability in CAIRNS’s output.>}
    \label{fig:scholarguide_prompt_visualization}
\end{figure}

CAIRNS then follows a ReAct-style reasoning-retrieval loop~\cite{yao2023react}. It retrieves evidence, generates an initial answer, and applies a sufficiency checker. If information gaps are found, the system generates new sub-questions and re-enters the retrieval loop to supplement the information until the answer is deemed sufficient. At the heart of answer generation in CAIRNS is the \textbf{ScholarGuide Prompt} (Figure~\ref{fig:scholarguide_prompt_visualization}). This prompt is central to balancing readability and factuality. For climate adaptation QA, the ScholarGuide prompt, which is a single structured prompt with guidance, explicitly informs the LLM of the grading rubric to optimise the answers. The prompt frames the LLM as an ``academic writing coach'' and adopts a three-way mechanism that factors (a) structure and clarity, (b) scientific reasoning and citations, and (c) specific and localized advice.



\paragraph{\bf Hybrid Evaluation Framework} 
We adopt the seven-dimensional human rubric of~\citet{Nguyen2025ClimateAdaptationQA} and augment it with faithfulness metrics. Firstly, we compute a faithfulness score by linearly combining an automatic metric (e.g., RAGAS) and an LLM-based assessor:
\begin{equation}
S_{\text{faithful}}=\alpha\,S_{\text{RAGAS}}+(1-\alpha)\,S_{\text{LLM}}, \quad \alpha\in[0,1].
\label{eq:hybrid-faithfulness}
\end{equation}
The weighting coefficient \( \alpha \) is validated on four external datasets to balance automatic and LLM assessments. 

To further align LLM-based scoring with expert judgments, we aggregate multiple evaluators using $\kappa$-based agreement weights. Note that these weights apply only to the seven rubric dimensions. For each evaluator $i$, Cohen’s $\kappa_i$ defines its normalized weight:
\begin{equation}
w_i=\frac{\max(\kappa_i,0)}{\sum_{j}\max(\kappa_j,0)}.
\label{eq:kappa-weight}
\end{equation}
The final hybrid score is the weighted sum
\begin{equation}
S_{\text{final}}=\sum_{i} w_i\,S_i,
\label{eq:final-score}
\end{equation}
where $S_i$ are component scores (including $S_{\text{faithful}}$ and other automated sub-scores). This $\kappa$-weighted fusion preserves interpretability while privileging evaluators that empirically align with experts.

\section{Experiment Setup}
\label{sec:experiments}
\paragraph{\bf Datasets and Tasks} 
We evaluate all models on a state-of-the-art, 50-question dataset provided in~\cite{Nguyen2025ClimateAdaptationQA}. The questions are written by domain experts and cover diverse regions and topics. The retrieval corpus consists of a set of 13000 publicly available climate-related papers from Elsevier. We automatically extract and normalize geographic information (\textit{country/adm1}), where \textit{country} denotes the nation and \textit{adm1} corresponds to major cities (e.g., \textit{Australia/Sydney}), as metadata to support location-weighted retrieval. Climate data are from INDRA API~\footnote{\href{https://research.csiro.au/indra/;  MyClimateCoPilot2025}{https://research.csiro.au/indra/}} (e.g., precipitation, temperature, and crop-yield indicators), transformed into structured text snippets for use during generation.

\paragraph{\bf Baselines and Models} 
We use \texttt{Gemini-2.0-Flash} in LLM-only baselines without external knowledge and only provide the original question. We experiment with multiple ablations of CAIRNS: (1) \textbf{Domain RAG (literature-only)}: retrieval from the literature subset only; (2) \textbf{Domain RAG (literature+data)}: retrieval augmented with both literature and climate data based on two-stage API reasoning. The base model in the case of CAIRNS and all its ablations is Gemini-2.0-Flash. All models share identical corpora, prompt structures, and evaluation settings for a fair evaluation.

\paragraph{\bf Evaluation Protocol} 
We build the Hybrid Evaluator using the \textit{My Climate CoPilot User Study Annotations} dataset~\cite{myclimatecopilot2025annotations}. Six commercial LLMs from three vendors are compared against human annotations, and Cohen's $\kappa$ scores are computed to measure agreement. These $\kappa$ values determine reliability weights $W_{m,d}$ for each model $m$ and evaluation dimension $d$. This calibrated Hybrid Evaluator is then used to score the outputs from our various CAIRNS configurations. We evaluate both zero-shot and few-shot settings, using five annotated examples per setting, also from~\cite{myclimatecopilot2025annotations}. The evaluator scores across seven dimensions: Context, Structure, Language, Comprehensiveness, Specificity, Citations, and Accuracy (range: 0-3, as in past work), and computes two additional faithfulness metrics (range: 0-1): Citation Rate and Faithfulness Rate.
\section{Results}
\label{sec:results}

\begin{table*}[tb]
\setlength{\tabcolsep}{4pt}
\centering
\caption{Comparative evaluation of CAIRNS and baseline pipelines across seven dimensions and faithfulness metrics. MTR = Multi-Turn ReAct. SGP = ScholarGuide Prompt. CD = Climate Data. LWR = Location-Weighted Retrieval. Bold values indicate the best score in each dimension. For ablation tests,\textbf{-} indicates that the component of CAIRNS that has been removed, and \textbf{+} indicates the external component that has been added. N/A: Not Applicable.}
\label{tab:overall_results}
\resizebox{\textwidth}{!}{
\begin{tabular}{lcccccccccccc}
\toprule
\textbf{Method} & \textbf{Context~$\uparrow$} & \textbf{Structure~$\uparrow$} & \textbf{Language~$\uparrow$} & \textbf{Citations~$\uparrow$} &
\textbf{Specificity~$\uparrow$} & \textbf{Compreh.~$\uparrow$} & \textbf{Accuracy~$\uparrow$} & \textbf{AVG~$\uparrow$} & \textbf{CitRate~$\uparrow$} & \textbf{FaithRate~$\uparrow$} \\
 & \textbf{(0-3)} & \textbf{(0-3)} & \textbf{(0-3)} & \textbf{(0-3)} &
\textbf{(0-3)} & \textbf{(0-3)} & \textbf{(0-3)} & \textbf{(0-3)} & \textbf{(0-1)} & \textbf{(0-1)} \\
\midrule
\multicolumn{11}{c}{\textbf{Baselines}} \\
\midrule
\texttt{Gemini (Zero-shot)} & 2.84±0.01 & 2.0±0.01 & 3.0±0.0 & 0.42±0.06 & 1.04±0.06 & \textbf{2.60±0.21} & 1.84±0.05 & 1.96±0.024 & N/A & N/A \\
\texttt{Gemini (Few-shot)} & 2.86±0.05 & 1.98±0.01 & 3.0±0.0 & 0.36±0.06 & 1.02±0.05 & \textbf{2.56±0.18} & 1.84±0.08 & 1.95±0.021 & N/A  & N/A \\
\midrule

\multicolumn{11}{c}{\textbf{CAIRNS}} \\
\midrule

\texttt{CAIRNS (Zero-shot)} & \textbf{3.0±0.0} & \textbf{3.0±0.0} & 3.0±0.0 & \textbf{2.94±0.0} & \textbf{1.86±0.02} & 2.50±0.13 & 2.78±0.07 & \textbf{2.72±0.02} & 0.528 & 0.814 \\
\texttt{CAIRNS (Few-shot)} & \textbf{3.0±0.0} & \textbf{3.0±0.0} & 3.0±0.0 & 2.94±0.0 & \textbf{1.88±0.11} & 2.24±0.18 & 2.80±0.06 & \textbf{2.69±0.03} & 0.528 & 0.814 \\
\midrule

\multicolumn{11}{c}{\textbf{Ablations of CAIRNS (Zero-shot)}} \\
\midrule

\texttt{-SGP -MTR} & 1.66±0.04 & 2.44±0.02 & 3.0±0.0 & 2.86±0.01 & 1.52±0.10 & 1.98±0.09 & 2.82±0.04 & 2.33±0.02 & 0.733 & 0.689 \\
\texttt{-SGP} & 1.78±0.07 & 2.44±0.05 & 3.0±0.0 & \textbf{2.94±0.02} & 1.60±0.03 & 2.18±0.11 & \textbf{2.86±0.07} & 2.39±0.02 & 0.740 & 0.799 \\
\texttt{-SGP -MTR +CD} & 1.50±0.03 & 2.36±0.02 & 2.96±0.01 & 2.82±0.02 & 1.58±0.07 & 2.02±0.21 & 2.48±0.06 & 2.23±0.03 & 0.754 & 0.688 \\
\texttt{-SGP +CD} & 1.64±0.05 & 2.44±0.05 & 3.0±0.0 & 2.90±0.06 & 1.54±0.06 & 2.14±0.12 & 2.64±0.05 & 2.32±0.03 & \textbf{0.778} & \textbf{0.832} \\
\texttt{-LWR} & \textbf{3.0±0.0} & \textbf{3.0±0.0} & 3.0±0.0 & 2.92±0.03 & 1.78±0.02 & 2.30±0.27 & 2.82±0.06 & 2.69±0.04 & 0.511 & 0.804 \\
\texttt{-MTR} & \textbf{3.0±0.0} & \textbf{3.0±0.0} & 3.0±0.0 & 2.88±0.04 & 1.82±0.03 & 2.34±0.15 & 2.84±0.07 & 2.69±0.02 & 0.490 & 0.781 \\
\midrule
\multicolumn{11}{c}{\textbf{Ablations of CAIRNS (Few-Shot)}} \\
\midrule

\texttt{-SGP -MTR} & 1.64±0.06 & 2.42±0.04 & 3.0±0.0 & 2.88±0.03 & 1.54±0.04 & 1.96±0.11 & \textbf{2.84±0.05} & 2.33±0.02 & 0.733 & 0.689 \\
\texttt{-SGP} & 1.78±0.05 & 2.44±0.01 & 3.0±0.0 & \textbf{2.96±0.01} & 1.60±0.09 & 2.00±0.14 & \textbf{2.84±0.05} & 2.37±0.02 & 0.740 & 0.799 \\
\texttt{-SGP -MTR +CD} & 1.50±0.05 & 2.34±0.05 & 2.96±0.01 & 2.82±0.04 & 1.54±0.08 & 2.16±0.09 & 2.48±0.07 & 2.25±0.02 & 0.754 & 0.688 \\
\texttt{-SGP +CD} & 1.64±0.06 & 2.46±0.04 & 3.0±0.0 & 2.94±0.05 & 1.54±0.06 & 2.22±0.07 & 2.60±0.05 & 2.34±0.01 & \textbf{0.778} & \textbf{0.832} \\
\texttt{-LWR} & \textbf{3.0±0.0} & \textbf{3.0±0.0} & 3.0±0.0 & 2.92±0.04 & 1.80±0.05 & 2.40±0.13 & 2.82±0.08 & 2.69±0.02 & 0.511 & 0.804 \\
\texttt{-MTR} & \textbf{3.0±0.0} & \textbf{3.0±0.0} & 3.0±0.0 & 2.86±0.02 & 1.84±0.11 & 2.20±0.13 & \textbf{2.84±0.06} & 2.68±0.02 & 0.490 & 0.781 \\
\bottomrule
\end{tabular}
}
\end{table*}
Table~\ref{tab:overall_results} shows that CAIRNS provides the best average scores across all experiments, with key trends described as follows. 

\paragraph{\bf Impact of ScholarGuide Prompt (SGP)} The ScholarGuide Prompt improves readability-related dimensions, including contextual richness, structural coherence, and language fluency. Meanwhile, its citation constraint effectively corrects inconsistencies across cited sources. 
Although this constraint slightly decreases the overall citation rate, it improves faithfulness, as indicated by a higher FaithRate compared to the `-SGP' ablation. 
By enforcing unified reference numbering and grounding each factual statement to explicit evidence, SGP enables the model to generate clearer, more verifiable, and location-specific answers. 
This pattern remains stable across both zero-shot and few-shot settings, demonstrating that structured prompting is a key driver of CAIRNS’s performance.

\paragraph{\bf Impact of Multi-turn ReAct (MTR)} The iterative reasoning loop allows the model to refine its intermediate outputs by repeatedly aligning retrieved literature and climate data, leading to more complete and context-aware answers, as reflected by higher comprehensiveness scores. While MTR’s contribution is modest relative to SGP, it complements the structured generation process and improves robustness on multi-source or data-intensive questions.

\paragraph{\bf Location-Weighted Retrieval (LWR)}  
Ablating the LWR component leads to a slight drop in specificity, while other metrics remain largely unchanged. This indicates that LWR plays a positive role in enabling the model to generate region-sensitive answers, particularly in questions where geographic focus is essential.

\paragraph{\bf Effect of Data-augmented Retrieval} We investigated the role of structured climate data in retrieval-augmented generation. Although we hypothesized that integrating structured climate data would enhance the factual support and interpretability of generated answers, the results suggest that this capability remains at an early stage. As shown in~Table~\ref{tab:overall_results}, the \texttt{DATA\_RAG} (literature+data) pipeline underperforms the \texttt{RAG} pipeline (literature-only) across most metrics, with the decline most pronounced in accuracy.  This indicates that the current model is still limited in its ability to leverage structured climate data to substantiate or refine textual reasoning.


\section{Evaluating the LLM-based Evaluator}
\label{sec:results-stability}
\begin{figure}[tb]
    \centering
    \includegraphics[width=0.9\linewidth]{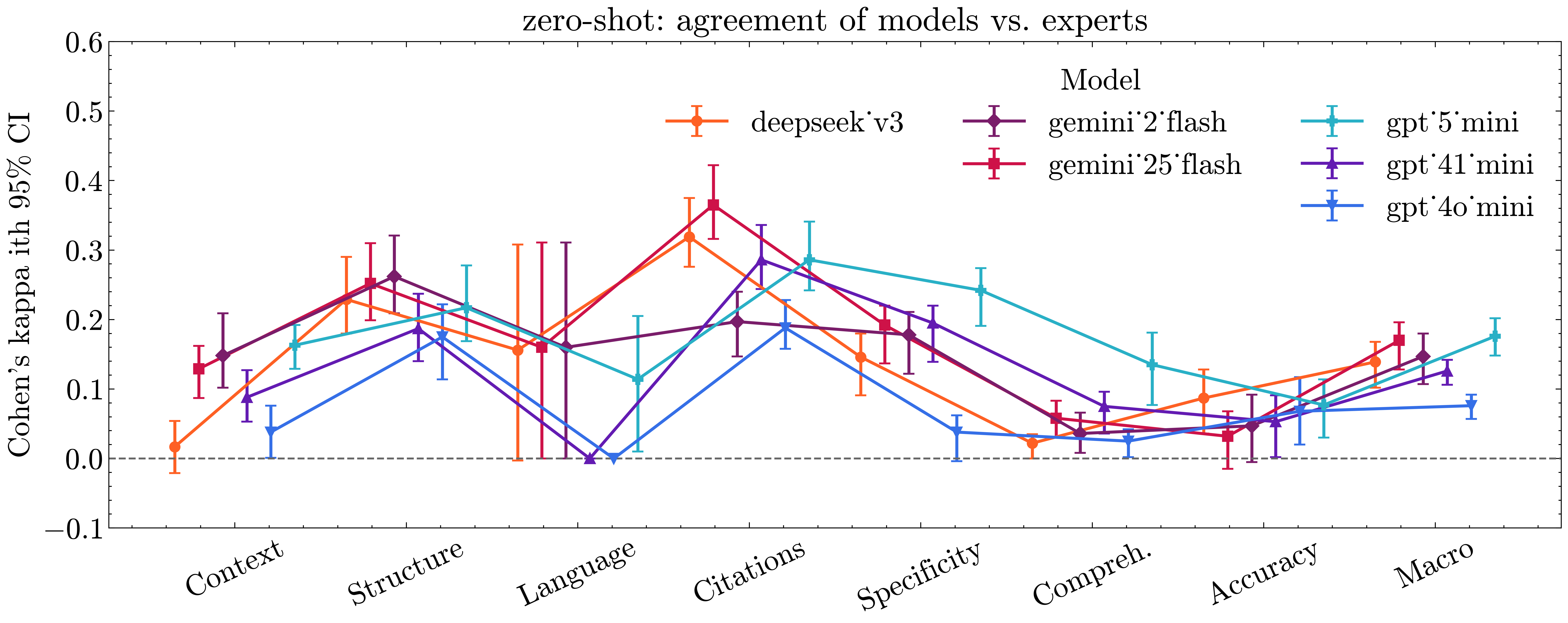}
    \caption{Zero-shot agreement between LLM evaluators and human experts across seven dimensions based on Cohen’s $\kappa$ with 95\% confidence interval (CI).}
    \Description[<A line chart showing the agreement (Cohen’s $\kappa$ with 95\% confidence intervals) between multiple large language models and human expert annotations in the zero-shot setting across seven evaluation dimensions.>]{<This figure illustrates the zero-shot agreement between various large language models (Gemini 2 Flash, Gemini 2.5 Flash, GPT-4o mini, GPT-4.1 mini, GPT-5 mini, and DeepSeek V3) and human expert annotations based on Cohen’s $\kappa$ metric with 95\% confidence intervals. The x-axis represents the seven evaluation dimensions—context, structure, language, citations, specificity, comprehensiveness, and accuracy—along with a macro average. The y-axis indicates $\kappa$ values ranging from -0.1 to 0.6, where higher values denote stronger agreement. The results show that while most models demonstrate moderate alignment in dimensions like structure and citations, the agreement declines in specificity and comprehensiveness. The wide confidence intervals in some cases reflect variability among model judgments in the absence of example guidance, highlighting the inherent instability of zero-shot evaluation.>}
    \label{fig:kappa_zero_shot}
\end{figure}

It is vital that the validity of LLM-based evaluation be scrutinized. To evaluate the evaluation framework itself, we first calculated the agreement between multiple LLM-based evaluators and human experts across all assessment dimensions using Cohen’s $\kappa$. This reflects the feasibility and trustworthiness of automated evaluation on each dimension. As illustrated in Figures~\ref{fig:kappa_zero_shot} and~\ref{fig:kappa_few_shot}, the consistency between LLM evaluators and experts varies notably across dimensions. For citations and structure, there is relatively high alignment, while specificity and comprehensiveness remain challenging, showing broader confidence intervals and lower $\kappa$ values. 
These differences reveal that individual evaluators demonstrate varying stability and reliability across different aspects of answer quality. As a result, we used the $\kappa$ values of each evaluator and dimension as weights to design a domain-specific Hybrid Evaluator tailored for agricultural climate adaptation QA. This evaluator integrates the strengths of multiple LLM-based assessors without requiring any fine-tuning or reinforcement learning, resulting in the highest overall consistency and credibility of automated evaluation observed in our study.

\begin{figure}[tb]
    \centering
    \includegraphics[width=0.9\linewidth]{Figures/kappa_few_shot.png}
    \caption{Few-shot agreement between LLM Evaluators and Human Experts across Seven Dimensions (Cohen’s $\kappa$ with 95\% CI).}
    \Description[<A line chart showing the agreement (Cohen’s $\kappa$ with 95\% confidence intervals) between multiple large language models and human expert annotations in the few-shot setting across seven evaluation dimensions.>]{<This figure depicts the agreement between multiple large language models (Gemini 2 Flash, Gemini 2.5 Flash, GPT-4o mini, GPT-4.1 mini, GPT-5 mini, and DeepSeek V3) and human experts in the few-shot evaluation setting. Cohen’s $\kappa$ values (with 95\% confidence intervals) are plotted across seven evaluation dimensions: context, structure, language, citations, specificity, comprehensiveness, and accuracy, along with a macro average. Compared to the zero-shot setting, the few-shot setting shows narrower confidence intervals and slightly higher $\kappa$ scores in most dimensions, indicating improved inter-model and model–expert consistency when provided with five annotated examples. The improvement is most evident in structure and citation dimensions, suggesting that few-shot exemplars help align LLM evaluators with human judgment standards for clear structure and reliable citation usage.>}
    \label{fig:kappa_few_shot}
\end{figure}

\section{Conclusions and Future Work}
\label{sec:conclusion}
\textbf{CAIRNS} is a question-answering framework for agricultural climate adaptation that balances readability, accuracy, and citation transparency.  It integrates structured prompting (SGP), multi-turn reasoning, and a consistency-weighted Hybrid Evaluator. Experimental results demonstrate that SGP substantially improves structural clarity and citation reliability without sacrificing factual precision, reflecting a meaningful step toward aligning QA outputs with domain researchers’ communication preferences. In contrast, data-augmented retrieval remains less effective at the current stage. By leveraging inter-model agreement as a weighting signal, our Hybrid Evaluator achieves the highest credibility among automated evaluations without fine-tuning or reinforcement learning. The evaluator leverages inter-model agreement as a weighting signal and can be directly applied to low-resource or expert-scarce settings for question-answering. In the future, we plan to enhance location-aware reasoning by incorporating fine-grained geographic information, such as latitude and longitude, to enable document transferability across similar climatic regions and to generalize literature-based insights. In parallel, we aim to establish a larger human-aligned evaluation benchmark that spans diverse question types and subdomains, serving both as a calibration resource for the Hybrid Evaluator and as a fine-tuning corpus for future domain-specific models, promoting reproducible and trustworthy QA research in climate adaptation.
\bibliographystyle{ACM-Reference-Format}
\balance
\bibliography{sample-base}
\end{document}